\pdfoutput=1

\documentclass[11pt]{article}

\usepackage[]{acl}

\usepackage{times}
\usepackage{latexsym}

\usepackage[T1]{fontenc}

\usepackage[utf8]{inputenc}

\usepackage{microtype}

\usepackage{graphicx}
\usepackage{setspace}
\usepackage{CJKutf8}
\usepackage{amsthm,amsmath,amssymb}
\usepackage{mathrsfs}
\usepackage{array}
\usepackage{colortbl}
\usepackage{arydshln}
\usepackage[ruled]{algorithm2e}
\usepackage{multirow}
\usepackage{color,xcolor}
\usepackage{bbding}
\usepackage[switch]{lineno}
\usepackage{footmisc} 
\usepackage{subfigure}

\title{Delving Deep into Regularity: A Simple but Effective Method for \\ Chinese Named Entity Recognition}

\author{{Yingjie Gu$^{1}$}, {Xiaoye Qu$^{1}$\thanks{\hspace*{1mm} Equal contribution}}, {Zhefeng Wang$^{1}$\thanks{\hspace*{1mm} Corresponding author}}, {Yi Zheng$^{1}$}, {Baoxing Huai$^{1}$}, {Nicholas Jing Yuan$^{1}$}  \\ 
  $^{1}$Huawei Cloud, China \\
  \tt \{guyingjie4,quxiaoye,wangzhefeng,zhengyi29,huaibaoxing\}@huawei.com \\ \texttt{nicholas.jing.yuan@gmail.com}
}

\begin{document}
\maketitle
\begin{abstract}
Recent years have witnessed the improving performance of Chinese Named Entity Recognition (NER) from proposing new frameworks or incorporating word lexicons. 
However, the inner composition of entity mentions in character-level Chinese NER has been rarely studied. Actually, most mentions of regular types have strong name regularity. For example, entities end with indicator words such as “\begin{CJK}{UTF8}{gbsn}公司\end{CJK} (company) ” or “\begin{CJK}{UTF8}{gbsn}银行\end{CJK} (bank)” usually belong to organization. 
In this paper, we propose a simple but effective method for investigating the regularity of entity spans in Chinese NER, dubbed as \textbf{R}egularity-\textbf{I}nspired {r}e\textbf{CO}gnition \textbf{N}etwork (RICON).
Specifically, the proposed model consists of two branches: a regularity-aware module and a regularity-agnostic module. The regularity-aware module captures the internal regularity of each span for better entity type prediction, while the regularity-agnostic module is employed to locate the boundary of entities and relieve the excessive attention to span regularity. An orthogonality space is further constructed to encourage two modules to extract different aspects of regularity features. 
To verify the effectiveness of our method, we conduct extensive experiments on three benchmark datasets and a practical medical dataset. The experimental results show that our RICON significantly outperforms previous state-of-the-art methods, including various lexicon-based methods.
\end{abstract}

\section{Introduction}

Named entity recognition (NER) aims at identifying text spans pertaining to specific entity types. It plays an important role in many downstream tasks such as relation extraction \cite{cheng2021hacred}, entity linking \cite{gu2021read}, co-reference resolution \cite{clark-manning-2016-improving}, and knowledge graph \cite{2020surveyknowledge}. 
Due to the complex composition \cite{gui-etal-2019-lexicon}, character-level Chinese NER is more challenging compared to English NER.
As shown in Figure 1 (a), the middle character “\begin{CJK}{UTF8}{gbsn}流\end{CJK}” can
constitute words with the characters to both their
left and their right, such as  “\begin{CJK}{UTF8}{gbsn}河流\end{CJK} (River)” and “\begin{CJK}{UTF8}{gbsn}流经\end{CJK} (flows)”, leading to ambiguous character boundaries. 

\begin{figure}[tp]
\centering
\includegraphics[width=1.0\columnwidth]{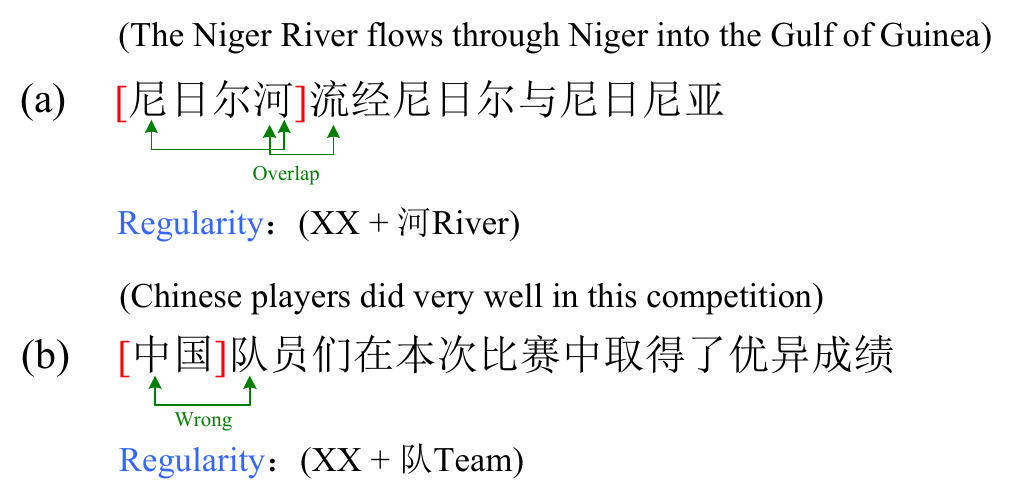}
\caption{(a) Complex composition of Chinese NER and regularity. (b) Excessive focusing on regularity leads to wrong entity boundary. }
\label{fig1}
\end{figure}

There are two typical frameworks for NER. 
The first one conceptualizes NER as a sequence labeling task \cite{2015Bidirectional,lample-etal-2016-neural,ma-hovy-2016-end}, where each character is assigned to a special label (e.g., B-LOC, I-LOC). 
The second one is span-based method \cite{2020Neural, yu-etal-2020-named}, which classifies candidate spans based on their span-level representations. 
However, despite the success of these two types of methods, they do not explicitly take the complex composition of Chinese NER into consideration. 
Recently, several works  \cite{zhang-yang-2018-chinese,gui-etal-2019-lexicon, li-etal-2020-flat} utilize external lexicon knowledge to help connect related characters and promote capturing the local composition. Nevertheless, building the lexicon is time-consuming and the quality of the lexicon may not be satisfied. 

In contrast to previous works, we observe that the regularity exists in the common NER types (e.g., ORG and LOC). As shown in Figure 1 (a), “\begin{CJK}{UTF8}{gbsn}尼日尔河\end{CJK} (Niger River)” 
follows the specific composition pattern “\begin{CJK}{UTF8}{gbsn}XX+河\end{CJK} (XX + River)” which ends with indicator character “\begin{CJK}{UTF8}{gbsn}河\end{CJK}" and mostly belongs to location type, and the ambiguous character “\begin{CJK}{UTF8}{gbsn}流\end{CJK}” can properly constitute “\begin{CJK}{UTF8}{gbsn}流经\end{CJK}” with the right character “\begin{CJK}{UTF8}{gbsn}经\end{CJK}”.
Thus, the regularity information serves as important clues for entity type recognition and identifying the character composition. 
Formally, we refer to regularity as specific internal patterns contained in a type of entity \cite{lin2020rigorous}. 
However, too immersed regularity leads to unfavorable boundary detection of entities and disturbing character composition. As shown in Figure 1 (b),
“\begin{CJK}{UTF8}{gbsn}中国队\end{CJK} (Chinese team)” conforms to the pattern “\begin{CJK}{UTF8}{gbsn}XX+队\end{CJK} (XX + Team)”, but the correct entity boundary should be “\begin{CJK}{UTF8}{gbsn}中国\end{CJK} (Chinese)” and “\begin{CJK}{UTF8}{gbsn}队员\end{CJK} (players)” according to the context. Therefore, the context also plays a key role in determining the character boundary.

In this paper, we introduce a simple but effective method to explore the regularity information of entity spans for Chinese NER, dubbed as \textbf{R}egularity-\textbf{I}nspired {r}e\textbf{CO}gnition \textbf{N}etwork (RICON). 
The proposed model   
consists of two branches named regularity-aware module and regularity-agnostic module, where each module has task-specific encoder and optimization object.  
Concretely, the regularity-aware module aims at analyzing the internal regularity of each span and integrates the significant regularity information into the corresponding span-level representation, leading to precise entity type prediction. 
Meanwhile, the regularity-agnostic module is devised to capture context information and avoid excessive focus on intra-span regularity.
Furthermore, we adopt an orthogonality space restriction to encourage two branches to extract different features with regard to the regularity. To verify the effectiveness of our method, we conduct extensive experiments on three  large-scale benchmark datasets (OntoNotes V4.0, OntoNotes V5.0, and MSRA). The results show that RICON achieves considerable improvements 
compared to the state-of-the-art models, even outperforming existing lexicon-based models. Moreover, we experiment on a practical medical dataset (CBLUE) to further demonstrate the ability of RICON. 

Our contributions can be summarized as follows: 
\begin{itemize}
\item This is the first work that explicitly explores the internal regularity of entity mentions for Chinese NER.  
\item We propose a simple but effective method for Chinese NER, which effectively utilizes regularity information while avoiding excessive focus on intra-span regularity.
\item Extensive experiments on three large-scale benchmark datasets and a practical medical dataset demonstrate the effectiveness of our proposed method. 
\end{itemize}

\section{Related Work}
Traditional methods treat NER as a sequence labeling task, where each word or character in the sentence is assigned to a special label. As
a representative,
\citet{2015Bidirectional} utilized the BiLSTM as an encoder to learn the contextual representation, and then exploited Conditional Random Field (CRF) as a decoder to label the tokens. The BiLSTM-CRF architecture achieved superior performance on various datasets, hence many following works  \cite{lample-etal-2016-neural,ma-hovy-2016-end} adopt such architecture. More recently, strong pre-trained language models such as ELMo \cite{peters-etal-2018-deep} and BERT \cite{devlin-etal-2019-bert} are incorporated to further enhance the performance of NER. Although the sequence labeling framework achieves decent performance on flat NER, it struggles for nested NER. As a result, span-based models are proposed to solve the nested problem by classifying all possible spans into predefined types (e.g. PER, LOC) in the sentence. For example,  
\citet{yu-etal-2020-named} adopted a biaffine attention model to assign scores for all potential spans and achieved the state-of-the-art performance on both flat and nested English NER datasets. \citet{shen2021locateandlabel} also employed span-based framework on Chinese NER datasets.
In this paper, we adopt span-based method as our basic framework for two reasons. Firstly, the span-based method considers each span and naturally suits analyzing inner-span character composition. Secondly, the span-based framework can easily extend our method from flat NER to nested NER.

\begin{figure*}
    \centering
    \includegraphics[width=16cm]{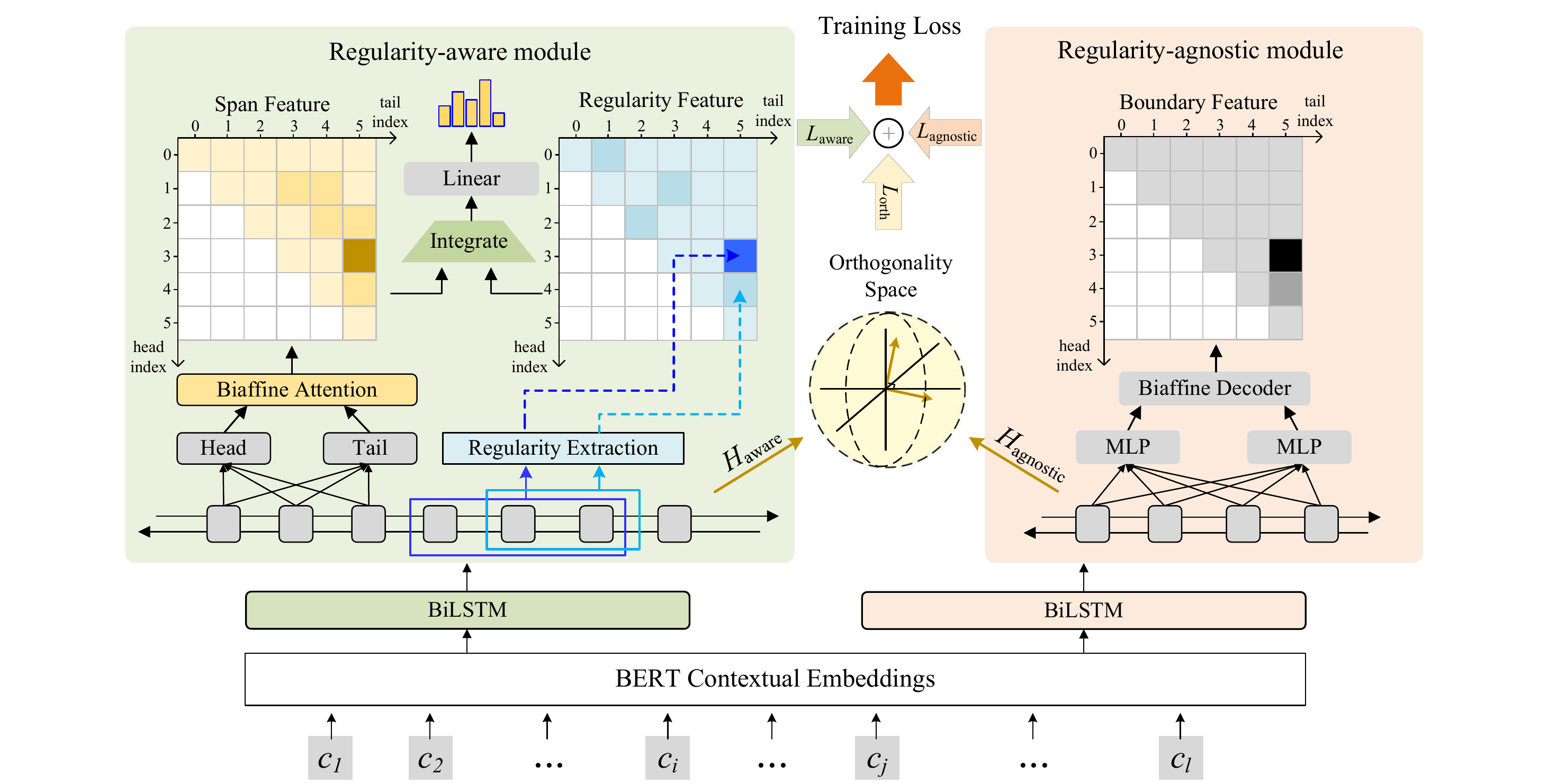}
    \caption{Overall structure of RICON. 
    Each character in the sentence is first embedded by BERT. Then, two separate Bi-LSTM layers are adopted to encode representations for the regularity-aware module and regularity-agnostic module. An orthogonality space is further utilized to encourage extracting different  features for each module.
    }
    \label{fig2}
\end{figure*}

Recently, for Chinese NER, researchers proposed various lexicon-based models that incorporate the external lexicon information and obtained better results. \citet{zhang-yang-2018-chinese} investigated Lattice-LSTM for incorporating word lexicons into the character-based NER model. However, the lattice structure fails to compute in parallel. To address this problem, \citet{gui-etal-2019-lexicon} introduced a lexicon-based graph neural network that recasts Chinese NER as a node classification task. There are also several works that focus on incorporating all matched words from the lexicon into the character embeddings \cite{ma-etal-2020-simplify, 2021Lexicon}. Different from the aforementioned lexicon-based works that incorporate external resources, in this paper, we focus on exploring the internal regularity information of spans.

\section{Method}

The overall architecture of our RICON is shown in Figure \ref{fig2}, which mainly consists of two branches: the regularity-aware module and the regularity-agnostic module. 

\subsection{Embedding and Task-specific Encoder}
First of all, each character of the input sequence is embedded into a dense vector. Then the character vectors are separately fed into two task-specific bidirectional LSTM (BiLSTM) layers to extract the corresponding hidden states for each module respectively.
Formally, given a sentence with $l$ characters $s=\{c_1,c_2,...,c_l\}$. We use a standard BERT
\cite{devlin-etal-2019-bert} to obtain the context dependent embeddings for a target token:
\begin{equation}
x_i = {\rm BERT}(c_i) 
\end{equation}
Then, the sequence of character embeddings will be fed to two separate BiLSTM layers for regularity-aware module and regularity-agnostic module. The hidden state of BiLSTM is expressed as follows:
\begin{equation}
\overrightarrow{h}_{i, \tau}=\overrightarrow{{\rm LSTM}}(x_i,\overrightarrow{h}_{i-1, \tau})
\end{equation}
\begin{equation}
\overleftarrow{h}_{i, \tau}=\overleftarrow{{\rm LSTM}}(x_i,\overleftarrow{h}_{i-1, \tau})
\end{equation}
\begin{equation}
{h}_{i, \tau}=[\overrightarrow{h}_{i, \tau};\overleftarrow{h}_{i, \tau}]
\end{equation}
where $\tau \in \{\text{\rm aware}, \text{\rm agnostic}\}$, [;] denotes concatenation, and the dimension of ${h}_{i, \tau}$ is $2d$. The character sequence representation can be denoted as ${H}_{\tau}=\{h_{1,\tau},...,h_{i,\tau},...,h_{l,\tau}\}$.

\begin{figure}[t]
\centering
\includegraphics[width=1.0\columnwidth]{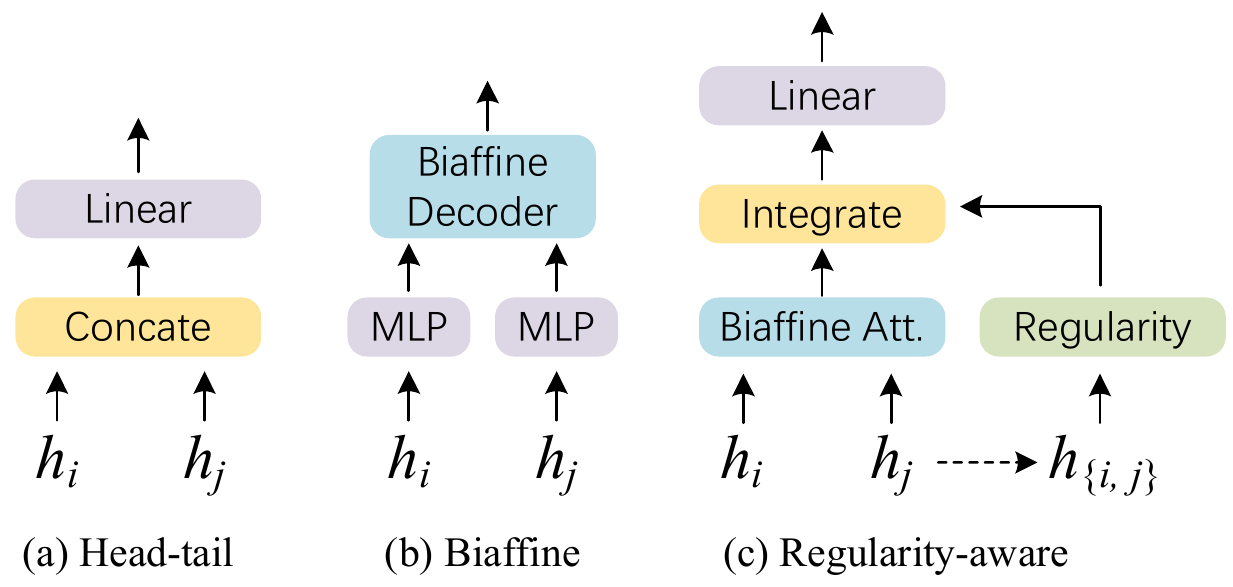}
\caption{Conceptual comparison of three architectures for span-based NER. $\{,\}$ denotes the representations across from $i\text{th}$ to $j\text{th}$ character of the span $s_{i,j}$.}
\label{fig3}
\end{figure}

\subsection{Regularity-aware Module}
In this module, we aim to explore the internal regularity of each span.  
As shown in Figure \ref{fig3} (a), typical span-based NER methods \cite{sohrab-miwa-2018-deep,xia-etal-2019-multi,2020Neural} represent each entity span via concatenating corresponding head and tail features, and use a linear classifier to predict the type of this span. In this way, the span features are coarse-grained. Then, as denoted in Figure \ref{fig3} (b), \citet{yu-etal-2020-named} propose a biaffine decoder to enhance the interaction between head and tail representations after two MLPs and predict span types simultaneously. 
Nevertheless, the internal regularity among characters in the span is still neglected in this biaffine method. 

Consequently for this, our regularity-aware module is devised to capture the internal regularity feature for each span $s_{i,j}$, as demonstrated in Figure \ref{fig3} (c). It is worth noting that span representations are obtained by the head and tail characters of the span, while the regularity representations stem from each character in the span.
To achieve this goal, we utilize a linear attention to obtain the regularity representation of each span as follows:
\begin{equation}
a_t=W_{\rm reg}^{\top} h_{t}+b_{{\rm reg}}
\end{equation}
\begin{equation}
\alpha_t=\frac{exp(a_t)}{\sum_{k=i}^jexp(a_k)}
\end{equation}
\begin{equation}
h_{s_{i,j}}^{({\rm reg})}=\sum_{t=i}^j\alpha_t \cdot h_{t}
\end{equation}
where $h_t = h_{t, \text{aware}}$ and $t\in \{i, i+1,..., j\}$ is the index of the span, $W_{\rm reg}\in\mathbb{R}^{2d\times 1}$ and $b_{\rm reg}\in\mathbb{R}^1$ are learnable weights and bias respectively. 
For a span whose length is 1, we do not extract extra features but use the hidden representation $h_{i,{\rm aware}}$ to denote its regularity. 
The regularity feature $H^{({\rm reg})}\in\mathbb{R}^{l\times l\times 2d}$ will be used for the subsequent entity type prediction.

To predict the type of an entity, our model integrates the regularity feature of each span into the span representations. Firstly, we acquire the span representation via a biaffine attention mechanism by interacting head and tail features:
\begin{equation}
h^{({\rm span})}_{s_{i,j}}={h}^{\top}_{i}U^{(1)}h_j + (h_i\oplus h_j)U^{(2)} + b_1
\end{equation}
where $h_i,h_j\in H_{\rm aware}$ are the head and tail representations of span $s_{i,j}$. $U^{(1)}$ is a ${2d\times 2d\times 2d}$ tensor, $U^{(2)}$ is a ${4d\times 2d}$ matrix, and $b_1$ is the bias.
It is worth noting that here we do not apply two separate MLPs like Figure 3 (b) to generate different representations for the head and tail features of the spans, as different MLPs will project the head, tail, and regularity representation into distinct spaces. The experiment also verifies that such space inconsistency degrades the recognition performance. 
Then a gated network is devised to integrate the span and regularity representation as below:
\begin{equation}
g_{s_{i,j}}=\sigma(U^{(3)}[h^{({\rm span})}_{s_{i,j}};h_{s_{i,j}}^{({\rm reg})}] + b_2)
\end{equation}
\begin{equation}
h_{s_{i,j}}=g_{s_{i,j}} \odot h^{({\rm span})}_{s_{i,j}} + (1-g_{s_{i,j}}) \odot h_{s_{i,j}}^{({\rm reg})}
\end{equation}
where $U^{(3)}\in \mathbb{R}^{4d\times 1}$ is a trainable parameter and $b_2$ is the bias. $\sigma$ denotes the sigmoid function and $\odot$ mean the element-wise dot multiplication. Finally, we adopt a standard linear classifier with a softmax function to predict the entity type for each span.
\begin{equation}
\tilde{y}_{s_{i,j}} = {\rm Softmax}({W_{\rm type}}^{\top}h_{s_{i,j}} + b_3)
\end{equation}
where $W_{\rm type}\in \mathbb{R}^{2d\times c}$ is a trainable parameter and $b_3$ is the bias.
The loss function of the regularity-aware module is defined as cross-entropy: 

\begin{equation}
    \mathcal{L}_{\rm aware} = -\frac{1}{N}{\sum^{N}_{n=1}\sum^l_{i=1}\sum^l_{j=1}}y^{(n)}_{s_{i,j}}{\rm log}(\tilde{y}^{(n)}_{s_{i,j}}), i\leq j
\end{equation}
where $\tilde{y}_{s_{i,j}}$ denotes the prediction and $y_{s_{i,j}}$ is the the ground truth type of the span. $N$ is the number of training samples in the regularity-aware module.

\subsection{Regularity-agnostic Module}
By considering regularity, above regularity-aware module makes the model stricter in terms of predicting the entity type, thus improving the precision of entity prediction. 
Nevertheless, too immersed regularity may result in inaccurate word boundaries.
To get rid of it, we propose to erase the concrete form of golden entities and relieve the excessive learning of structural pattern by regularity-aware module. 
In this scenario, the head and tail features which determine boundary become more significant, thereby we first apply two multi-layer perceptrons (MLPs) on the hidden states from BiLSTM to get separate representations for head and tail. 
Then a biaffine decoder is leveraged for obtaining entity probability of the span $s_{i,j}$ as follows:

\begin{equation}
\bar{h}_i={\rm MLP}_{\rm head}(h_{i}) \quad \bar{h}_j={\rm MLP}_{\rm tail}(h_{j})
\end{equation}


\begin{equation}
\bar{y}_{ij} = \sigma({[\bar{h}_i;1]}^{\top} U_m {[\bar{h}_j;1]})
\end{equation}
where $h_i = h_{i, \text{agnostic}}$, $h_j = h_{j, \text{agnostic}}$, $U_m$ is a $(2d+1)\times 1\times (2d+1)$ trainable parameter, $\sigma$ is the sigmoid function.
Finally, we adopt binary cross-entropy loss to train this task.
\begin{equation}
\begin{split}
    \mathcal{L}_{\rm agnostic}=-\frac{1}{N}{\sum^{N}_{n=1}\sum^l_{i=1}\sum^l_{j=1}}[y^{(n)}_{ij}{\rm log}(\bar{y}^{(n)}_{ij}) \\+
    (1-y^{(n)}_{ij}){\rm log}(1-\bar{y}^{(n)}_{ij})], i\leq j
\end{split}
\end{equation}
where $\bar{y}_{ij}$ denotes the prediction and $y_{ij}$ is the binary target indicating whether the span is an entity or not. $N$ is the number of training samples in the regularity-agnostic module. 

\subsection{Orthogonality Space Restriction}
As regularity-aware module aims to capture the regularity information while regularity-agnostic module pays no attention to the concrete regularity,
we expect to learn different features for these two modules. To this end, we construct an orthogonality space on the top of two BiLSTM layers to encourage encoding different aspects of the input embeddings. The loss is calculated as follows:

\begin{equation}
    H_{\rm orth} =  {H_{\rm aware}}^{\top}H_{\rm agnostic}
\end{equation}
\begin{equation}
    \mathcal{L}_{\rm orth} ={\lVert H_{\rm orth}\rVert}^2_F=-\frac{1}{N}\sum^{N}_{n=1}\sum_{i=1}^l\sum_{j=1}^l {\vert h^{(n)}_{ij} \vert}^2
\end{equation}
where ${\lVert \cdot \rVert}^2_F$ is the squared Frobenius norm and $N$ is the number of training elements.

\subsection{Training and Inference}
During training, our RICON can be trained by joint optimizing above three sub-tasks, so we define the total loss as below:
\begin{equation}
    \mathcal{L} = \lambda_1\mathcal{L}_{\rm aware} + \lambda_2\mathcal{L}_{\rm agnostic} + \lambda_3\mathcal{L}_{\rm orth}
\end{equation}
where $\lambda_1$, $\lambda_2$, and $\lambda_3$ are hyperparameter. 
During inference, 
we directly use regularity-aware module to predict the entity type for each span and apply a post-processing constraint for two overlapped entity candidates $E_1$ and $E_2$ that if ${E_1}_i < {E_2}_i \leq {E_1}_j < {E_2}_j$, where $i$ and $j$ are start and end indexes, we only select the entity with the higher type score.



\section{Experiments}

\subsection{Datasets}

\noindent\textbf{OntoNotes V4.0} \cite{0OntoNotes}. It is a  multilingual
corpus in the news domain. This dataset has 4 entity types. We use the same split as \cite{zhang-yang-2018-chinese}.

\noindent\textbf{OntoNotes V5.0} \cite{pradhan-etal-2013-towards}. Compared with V4.0, this version has more news data and contains 18 types of entities. We use the same split as \cite{2019Dependency}.

\noindent\textbf{MSRA} \cite{levow-2006-third}. It contains 3 types of named entities collected from the news domain. We use the same split as \cite{gui-etal-2019-lexicon}.

\noindent\textbf{CBLUE-CMeEE} \cite{hongying2020building}. CBLUE is Chinese biomedical language understanding evaluation which consists of 10 sub-tasks. Among them, CMeEE focuses on Chinese medical entity extraction and has 9 types of entities. We use the official train and dev split.

\noindent In addition, all types of OntoNotes V4.0, OntoNotes V5.0, MSRA, and 8 types of CBLUE-CMeEE are flat NER, while the symptom type of CBLUE-CMeEE is nested NER. 

Due to the space limitation, the statistics of all datasets are listed in the appendix.


\begin{table*}[t]\small
\centering
\scalebox{0.92}{
\setlength{\tabcolsep}{1.5mm}{
\begin{tabular}{lcccccccccc}
\hline
\multicolumn{1}{l|}{\multirow{2}{*}{Models}} & \multicolumn{1}{l|}{\multirow{2}{*}{Lexicon}} & \multicolumn{3}{c|}{OntoNotes V4.0} & \multicolumn{3}{c|}{OntoNotes V5.0} & \multicolumn{3}{c}{MSRA} \\ \cline{3-11} 
\multicolumn{1}{l|}{}                        & \multicolumn{1}{l|}{}                         & P   & R  & \multicolumn{1}{c|}{F1}  & P   & R  & \multicolumn{1}{c|}{F1}  & P      & R      & F1     \\ \hline\hline
\multicolumn{1}{l|}{Lattice LSTM \cite{zhang-yang-2018-chinese}}                       & \multicolumn{1}{c|}{\checkmark}           &        76.35     & 71.56       & \multicolumn{1}{c|}{73.88}    &  -   &  -  & \multicolumn{1}{c|}{-}    &       93.57  & 92.79  & 93.18     \\
\multicolumn{1}{l|}{Collaborative Graph Network \cite{sui-etal-2019-leverage}}                       & \multicolumn{1}{c|}{\checkmark}                         &    75.06         & 74.52    & \multicolumn{1}{c|}{74.79}    &  -   &  -  & \multicolumn{1}{c|}{-}    &   94.01      & 92.93      & 93.47    \\
\multicolumn{1}{l|}{LGN \cite{gui-etal-2019-lexicon}}                       & \multicolumn{1}{c|}{\checkmark}                         &     76.13     & 73.68   & \multicolumn{1}{c|}{74.89}    &    - &  -  & \multicolumn{1}{c|}{-}    &      94.19  & 92.73  & 93.46       \\
\multicolumn{1}{l|}{DGLSTM-CRF \cite{2019Dependency}}                       & \multicolumn{1}{c|}{}                         &  -   &  -  & \multicolumn{1}{c|}{-}    &   77.40     & 77.41    & \multicolumn{1}{c|}{77.40}    &     -   &    -    &    -    \\
\multicolumn{1}{l|}{WC-GCN \cite{2020Word}}                       & \multicolumn{1}{c|}{\checkmark}                         &  76.59   &  75.17  & \multicolumn{1}{c|}{75.87}    &   -  & -   & \multicolumn{1}{c|}{-}    &    94.82    &    93.98    &    94.40    \\
\multicolumn{1}{l|}{Star-GAT \cite{chen-kong-2021-enhancing} }                       & \multicolumn{1}{c|}{}                         &  79.25   & 80.66   & \multicolumn{1}{c|}{79.95}    &   78.22  &   80.88 & \multicolumn{1}{c|}{79.53}    &     -   &    -    &   -     \\ \hline
\textbf{with Pre-trained Language Model}              &                                               &     &    &                          &     &    &                          &        &        &        \\ \hline
\multicolumn{1}{l|}{BERT-Tagger}                        & \multicolumn{1}{c|}{}                         &  76.01   &  79.96  & \multicolumn{1}{c|}{77.93}    &   73.59  &  80.55  & \multicolumn{1}{c|}{76.91}    &    93.40    &   94.12     &   93.76     \\
\multicolumn{1}{l|}{BERT+LSTM+CRF}                        & \multicolumn{1}{c|}{}                         &   81.99  &   81.65 &  \multicolumn{1}{c|}{81.82}    &  77.12   &  79.81  & \multicolumn{1}{c|}{78.44}    &     95.06  & 94.61  & 94.83       \\
\multicolumn{1}{l|}{BERT+PLTE \cite{mengge-etal-2020-porous}}                        & \multicolumn{1}{c|}{\checkmark}                         &  79.62     & 81.82    & \multicolumn{1}{c|}{80.60}    &  -     & -  & \multicolumn{1}{c|}{-}    &    94.91  & 94.15  & 94.53     \\
\multicolumn{1}{l|}{BERT+Biaffine \cite{yu-etal-2020-named}}                        & \multicolumn{1}{c|}{}                         &  81.06     & 84.03    & \multicolumn{1}{c|}{82.52}    &  78.79     & 80.07  & \multicolumn{1}{c|}{79.43}    &    \textbf{96.65}  & 94.75  & 95.20     \\
\multicolumn{1}{l|}{BERT+FLAT \cite{li-etal-2020-flat}}                        & \multicolumn{1}{c|}{\checkmark}                         &  -   &  -  & \multicolumn{1}{c|}{81.82}    &   -  &  -  & \multicolumn{1}{c|}{-}    &    -    &     -   &    96.09    \\
\multicolumn{1}{l|}{BERT+SoftLexicon \cite{ma-etal-2020-simplify}}                        & \multicolumn{1}{c|}{\checkmark}                         &    \textbf{83.41}     & 82.21  & \multicolumn{1}{c|}{82.81}    &  -   &  -  & \multicolumn{1}{c|}{-}    &      95.75      & 95.10      & 95.42    \\
\multicolumn{1}{l|}{LEBERT \cite{2021Lexicon}}                        & \multicolumn{1}{c|}{\checkmark}                         &  -   &  -  & \multicolumn{1}{c|}{82.08}    &   -  &  -  & \multicolumn{1}{c|}{-}    &    -    &     -   &    95.70    \\ \hline
\multicolumn{1}{l|}{RICON (Ours)}                        & \multicolumn{1}{c|}{}                         &   81.95     & \textbf{84.78}   & \multicolumn{1}{c|}{\textbf{83.33}}    &   \textbf{79.26}     & \textbf{81.64}   & \multicolumn{1}{c|}{\textbf{80.43}}    &      95.94      & \textbf{96.33}     & \textbf{96.14} \\ \hline
\end{tabular}}}
\caption{We compare our RICON with recent state-of-the-art models on three Chinese benchmark datasets.}
\label{Tab2}
\vspace{-2mm}
\end{table*}

\subsection{Implementation Details}
In our experiments, we use the same settings for all datasets. Specifically, we adopt the standard pre-trained Chinese BERT-base model with 768 dimensions hidden representation to obtain character embeddings. We use Adam optimizer with 2e-5 learning rate for BERT embedding fine-tuning and 0.001 learning rate for other parts. The number of layer and dropout rate of BiLSTM encoders are set to 3 and 0.4. The hidden state size of BiLSTM encoders is set to 200. 
For the regularity-agnostic module, the output dimension of MLPs and the dropout rate are set to 150 and 0.2. To avoid overfitting, we also apply 0.1 dropout rate for the BERT output embeddings. For the hyper-parameters in loss, we set $\lambda_1=\lambda_2=1$ and $\lambda_3=0.5$. For all experiments including ablation study, we adopt an average of performance over five different runs to reduce randomness.

\subsection{Comparison Methods}
In our experiments, we compare our RICON with 
recent state-of-the-art methods, where part of them contain pre-trained language model BERT or external Chinese lexicon information. Here we briefly describe five typical methods:



\noindent

\noindent (1) \textbf{Star-GAT} \cite{chen-kong-2021-enhancing} propose a Star-transformer based NER system. They utilize explicit head and tail boundary information and Dependency GAT-based implicit boundary information to improve the performance. It is the SOTA model on the OntoNotes V5.0 dataset.

\noindent (2) \textbf{BERT+Biaffine} \cite{yu-etal-2020-named} recast NER as a task of identifying start and end positions and assigning a type to each span by a biaffine attention.

\noindent (3) \textbf{BERT+FLAT} \cite{li-etal-2020-flat} devise a FLAT model for Chinese NER, which converts the lattice structure into a flat structure consisting of spans to overcome the shortage of lattice-based model \cite{zhang-yang-2018-chinese}. They also equipped with BERT embeddings and achieved the SOTA performance on the MSRA dataset.

\noindent (4) \textbf{BERT+SoftLexicon} \cite{ma-etal-2020-simplify} incorporate the word lexicon into the character features. They leverage Chinese lexicon to match every character in the sentence with word appeared in the lexicon to improve the performance, which achieves the SOTA performance on OntoNotes V4.0.

\noindent (5) \textbf{LEBERT} \cite{2021Lexicon} introduce a Lexicon Adapter layer to integrate external lexicon knowledge into BERT layers directly. 

\subsection{Results}
We present the results on three benchmark datasets in Table \ref{Tab2}. From this table, we can observe that our RICON achieves the state-of-the-art performance on these datasets. Moreover, RICON even outperforms recent methods with Chinese lexicon significantly. Concretely, on OntoNotes V4.0, RICON achieves 0.81 absolute F1 improvement over the strong method BERT+Biaffine and 0.52 absolute improvement compared with the SOTA lexicon-based method BERT+SoftLexicon. On OntoNotes V5.0, we obtain a decent improvement compared to the SOTA approach Star-GAT by 0.90 F1 score. In addition, on MSRA, although the improvement of our model over the SOTA model BERT-FLAT is limited, our model still surpasses the other two lexicon-based models LEBERT and BERT+SoftLexicon by 0.44 and 0.72 respectively.

In addition, we present the model performance on CBLUE-CMeEE in Table \ref{Tabm}. Considering there are no available lexicons for this task, we only compare RICON with typical models. As shown in this table, RICON outperforms the strong BERT-Biaffine model with a 3.28 F1 score improvement over 9 types. It is remarkable progress in this challenging dataset. Meanwhile, we provided the result of nested symptom type. RICON performs much better than BERT-Biaffine with a 5.81 F1 improvement. This observation also denotes that our RICON also applies to nested NER. 

\begin{table}[t]\small
\centering
\setlength{\tabcolsep}{1.1mm}{
\begin{tabular}{lllllll}\hline
\multicolumn{1}{c}{} & \multicolumn{3}{c}{All Types} & \multicolumn{3}{c}{Symptom Type} \\ \hline\hline
Model                & P        & R         & F1       & P          & R        & F1       \\ \hline
BERT-Tagger    &   53.41    &     63.32      &    57.95       &   40.57       &    45.38          &  42.84                \\
BERT-CRF             & 58.34     & 64.08     & 61.07    & 46.01      & \textbf{47.51}     & 46.75     \\
BERT-Biaffine        & 64.17     & 61.29     & 62.29    & \textbf{63.17}      & 33.91     & 44.14     \\
RICON                & \textbf{66.25}     & \textbf{64.89}     & \textbf{65.57}    & 57.93      & 43.99     & \textbf{50.01}    \\\hline

\end{tabular}}
\caption{Performance of models on CBLUE-CMeEE, including all types and symptom type.} 
\label{Tabm}
\end{table}

\subsection{Ablation Study}

We conduct abundant ablation studies on OntoNotes V4.0 and V5.0 from module and implementation perspectives in Table \ref{Tab3} and \ref{Tab4}. Vanilla in tables is built from RICON by removing orthogonality space and regularity-agnostic module, and omitting to capture regularity features and integrate it in the regularity-aware module.

From the results in Table \ref{Tab3}, we can observe that: 
(1) When applying regularity-agnostic module to the vanilla, the performances improve by 0.21 and 0.48 respectively, showing the effectiveness of this module.
(2) When the vanilla equips with regularity-aware module, the F1 scores significantly improve by 0.57 and 0.65 respectively, which verifies that regularity plays a significant role in entity recognition.
(3) After combining regularity-aware and regularity-agnostic modules, we achieve further improvements, which indicates that two modules can mutually reinforce each other.
(4) The orthogonality space is a valid method according to the further F1 score improvements.

\begin{table}[t]\small
\centering
\scalebox{0.92}{
\setlength{\tabcolsep}{0.6mm}{
\begin{tabular}{l|ccc|ccc}
\hline
 \multirow{2}{*}{Module} & \multicolumn{3}{c|}{OntoNotes V4.0} & \multicolumn{3}{c}{OntoNotes V5.0} \\ \cline{2-7} 
                                                 & P          & R          & F1        & P          & R         & F1        \\ \hline \hline
Vanilla       & 81.08      & 84.17      & 82.59     & 77.87      & 82.04     & 79.42    \\ \hline 
+Reg-agnostic      & 81.18      & 84.77      & 82.80     & 78.32      & 81.48     & 79.90   \\ \hline 
+Reg-aware       & 82.49      & 83.86      & 83.16     & 79.28      & 80.90     & 80.07     \\ \hline
+Reg-aware \& agnostic        & 81.72      & 84.89      & 83.28     & 79.24      & 81.48     & 80.33     \\ \hline
RICON (Ours)             & 81.95      & 84.78      & 83.33     & 79.26      & 81.64     & 80.43     \\ \hline

\end{tabular}}}
\caption{Performance of modules on OntoNotes.} 
\label{Tab3}
\end{table}

Furthermore, we notice that adding the regularity-aware module significantly increases the \textbf{Precision} (1.41 on both datasets, Vanilla vs Vanilla+Reg-aware) but reduces the \textbf{Recall} (0.31 and 1.04 respectively), which conforms to that focusing on regularity feature would reinforce the type prediction, while missing several spans that are supposed to be entities.
Nevertheless, this situation can be remedied by the regularity-agnostic module and the \textbf{Recall} improved 1.03 and 0.58, respectively (Vanilla+Reg-aware vs Vanilla+Reg-aware \& agnostic). This result also meets our motivation that regularity-agnostic module can reinforce the entity boundary detection.

As shown in Table \ref{Tab4}, there are several alternative ways to extract regularity information instead of linear attention used in this paper, such as mean-pooling, max-pooling, or more complex multi-head self-attention \cite{vaswani2017attention}, but these methods all perform worse. It is one future direction to explore how to obtain regularity by a more sophisticated architecture. However, considering the model complexity and performance, we choose linear attention to capture regularity. 
In addition, replacing our devised gate mechanism with a simple concatenate or add operation both degrades the performance, denoting that gate mechanism is more efficient to integrate span feature and regularity feature. 
We also explored adding two MLPs separately to head and tail features when generating span features in the regularity-aware module. The experimental results prove that different feature space for span feature and regularity feature leads to worse performance. 

\begin{table}[t]\small
\centering
\scalebox{0.85}{
\setlength{\tabcolsep}{1.2mm}{
\begin{tabular}{l|c|c}
\hline
\multirow{2}{*}{Implementation} & \multicolumn{2}{c}{Dataset (F1)} \\ \cline{2-3} 
                                             & OntoNotes  V4.0          & OntoNotes V5.0          \\ \hline \hline
Vanilla+Reg-aware            & 83.16         & 80.07         \\ \hline 
Reg. feature by Mean-pooling            & 83.06 (-{0.10})         & 79.97 (-{0.10})         \\
Reg. feature by Max-pooling             & 82.82 (-{0.34})         & 79.79 (-{0.28})         \\ 
Reg. feature by Multi-Head   & 83.10 (-{0.06})         & 79.86 (-{0.21})         \\ \hline
Gate replaced with Add  & 82.96 (-{0.20})         & 79.94 (-{0.13})         \\
Gate replaced with Cat  & 82.80 (-{0.36})         & 79.77 (-{0.30})         \\ \hline
Apply MLPs to head and tail & 82.90 (-{0.26})         & 79.67 (-{0.40})         \\ \hline
Vanilla            & 82.59 (-{0.57})         & 79.42 (-{0.65})         \\\hline    
\end{tabular}}}
\caption{Performance of variants on OntoNotes datasets.} 
\label{Tab4}
\end{table}

\begin{figure*}[tp]
\centering
\includegraphics[width=1.9\columnwidth]{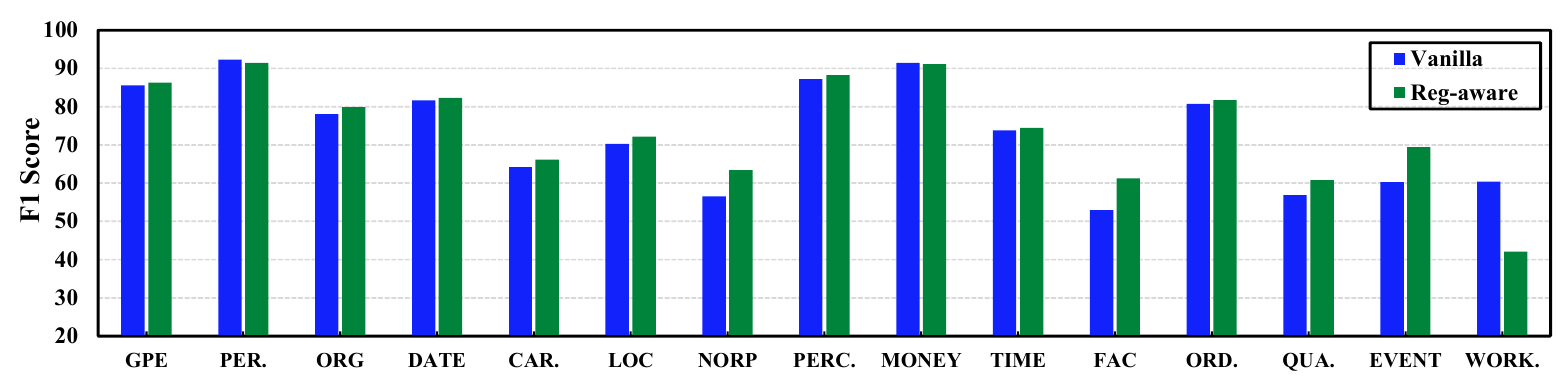}
\caption{The performance of 15 types of entities on OntoNotes V5.0. The types are sorted in descending order based on the proportion of entities of that type to the total. As the remaining 3 types on OntoNotes V5.0 only have less than 35 entities (0.05\%) among all entities. To avoid the impact of labeling errors, we do not present them here.}
\label{fig4}
\end{figure*}

\subsection{Analysis} 
In this section, We deeply analyze our proposed RICON from the following aspects.

\begin{figure}[tp]
  \centering
  \subfigure[F1 on OntoNotes V4.0]{\includegraphics[width=0.48\columnwidth]{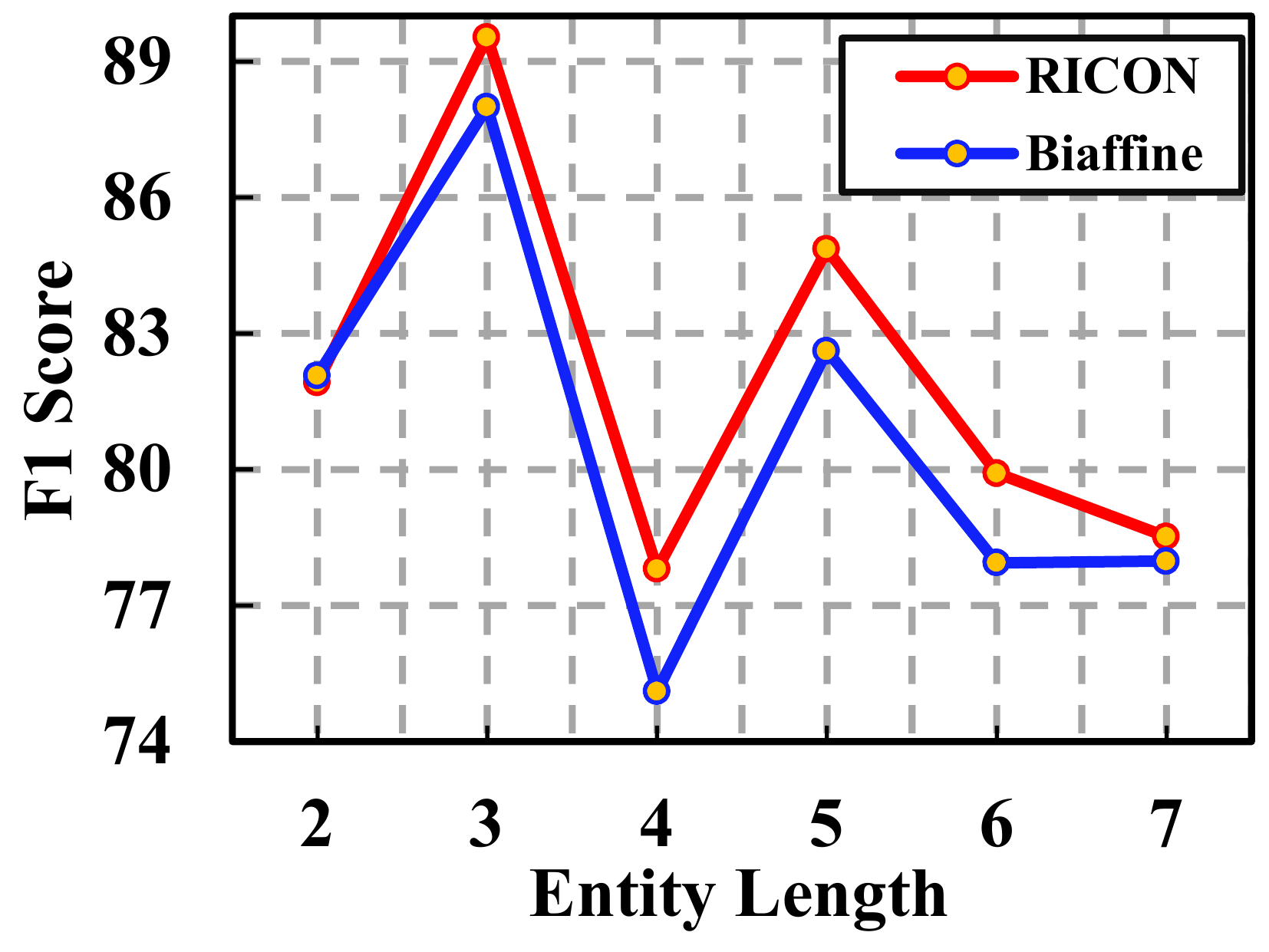}}
  \subfigure[F1 on OntoNotes V5.0]{\includegraphics[width=0.48\columnwidth]{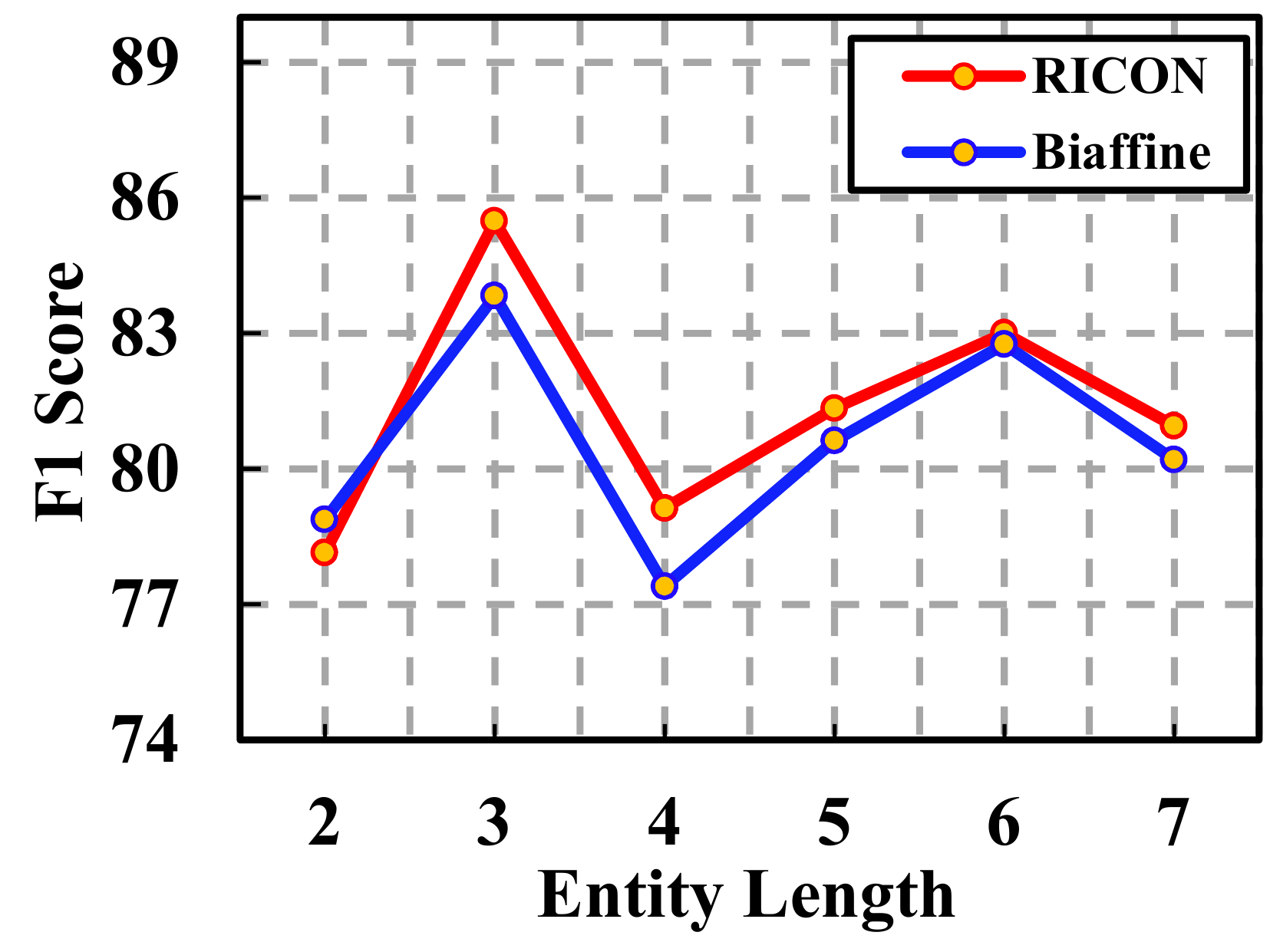}}
  \caption{Performance vs. Entity Length}
  \label{fig5}
  \vspace{-4mm}
\end{figure}

\begin{CJK}{UTF8}{gbsn}
\begin{table*}[t]\small
\centering
\begin{tabular}{|l|ccccccc|}
\hline
\#1 Sentence (Truncated)         & \multicolumn{7}{l|}{\begin{tabular}[c]{@{}l@{}}据 报 道， 从 波 罗 的 \textcolor{orange}{海} 三 国 撤 回 的 俄 罗 斯 军 队......  \\ (Reportedly, Russian Army withdrawn from the three countries around Baltic \textcolor{orange}{Sea}...)\end{tabular}}                                                                                                             \\
Characters (Entity Included) & 波                   & 罗                   & 的                   & \textcolor{orange}{海}                   &                     &                     &                     \\
Gold Label           & B-LOC               & M-LOC               & M-LOC               & E-LOC               &                     &                     &                     \\
Vanilla              & B-GPE               & M-GPE               & M-GPE               & M-GPE               &                     &                     &                     \\
Vanilla + Reg-aware             & B-LOC            & M-LOC              & M-LOC             & E-LOC             &                     &                     &                     \\
Regularity weight            & 0.04               & 0.06               & 0.07               & \textbf{0.83}               &                     &                     &                     \\
\hline

\hline

         \#2  Sentence (Truncated)         & \multicolumn{7}{l|}{\begin{tabular}[c]{@{}l@{}}新 闻 分 析 ： 美 国\textcolor{blue}{ 公 司} 兼 并 为 何 愈 演 愈 烈?\\ (News analysis: why the mergers of American \textcolor{blue}{companies} are intensifying?)\end{tabular}} \\
Characters (Entity Included) & 美                   & 国                   & \textcolor{blue}{公}                   & \textcolor{blue}{司}                   &                     &                     &                     \\
Gold Label            & B-GPE               & E-GPE               & O              & O               &                     &                     &                     \\
Vanilla + Reg-aware           & B-ORG               & M-ORG               & M-ORG               & M-ORG               &                     &                     &                     \\
Reg-aware + Reg-agnostic             & B-GPE               & E-GPE               & O               & O               &                     &                     &                     \\

\hline                                           
\end{tabular}
\caption{There examples from the Ontonotes V4.0 dataset. The label is organized in the form of BMES.}
\label{Tab5}
\vspace{-4mm}
\end{table*}
\end{CJK}




\subsubsection{Regularity: A Latent Adaptive Lexicon.}

The lexicon-based methods focus on incorporating external word lexicons to improve the performance of character-based NER. The core concept of them is preserving all words which match a specific character and let the subsequent NER model determine which word to apply \cite{zhang-yang-2018-chinese, ma-etal-2020-simplify}. In our model, we calculate the regularity for each span, namely, all words containing a specific character are considered, and then the best word and corresponding regularity will be determined. In this sense, our explored regularity can be seen as a latent adaptive lexicon. 
Furthermore, this latent adaptive lexicon is more complete than external lexicons because all spans matching the specific character are considered, while lexicon-based methods only match a limited number of words.
As shown in Table \ref{Tab2}, the previous SOTA method BERT+Biaffine performs worse than lexicon-based methods, but our regularity-based method RICON outperforms the lexicon-based methods. Actually, our regularity-based method can further be combined with lexicon-based methods.


\subsubsection{Performance vs. Entity Type.}
We examine how regularity affects each entity type.
As Figure \ref{fig4} shows, 12 types of entities achieve better performance with the regularity. This result conforms to the fact that types like GPE, ORG, and DATE have strong regularity. 
Nevertheless, for the types with little regularity information, such as WORK\_OF\_ART and PERSON, immersed regularity leads to performance degradation. We notice that the MONEY type typically contains regularity but we do not observe an improvement in this category. This is, due to inconsistencies between the training and test dataset. For instance, the training data contains the abundant pattern "number+dollar", while only numbers exist in the test set.
To remedy the excessive regularity, our RICON further utilizes a regularity-agnostic module to rectify the captured regularity. 
The above observations also inspire us to devise more elaborate NER for different entity types with various degree regularity properties in the future. Our regularity-aware module may also serve as a potential tool for evaluating the intensity of regularity. 


\subsubsection{Performance vs. Entity Length.}
Figure \ref{fig5} depicts the performance on the OntoNotes V4.0 and V5.0 datasets with different length of entities. From this figure, we can observe that our RICON consistently outperforms BERT-Biaffine \cite{yu-etal-2020-named} when the entity length is longer than 2, which illustrates that the regularity information is helpful to predict the types for long entities. 
In contrast, BERT-Biaffine performs comparable to RICON when entity length is 2 as there are no additional character information except the head and tail representations.

\subsubsection{Case study.}
Table \ref{Tab5} shows two examples from OntoNotes V4.0. 
In the first example, the Vanilla misidentifies the entity type, while Vanilla+reg-aware learns regularity “\begin{CJK}{UTF8}{gbsn}XX+海\end{CJK}"  by the greatest weight 0.83 on “\begin{CJK}{UTF8}{gbsn}海\end{CJK}", thus obtaining the accurate entity type. It is worth noting that regularity can capture more complex character compositions besides explicit patterns in the first example. More complex examples are presented in the appendix. 
In the second example, “\begin{CJK}{UTF8}{gbsn}美国公司\end{CJK}" conforms to the regularity "\begin{CJK}{UTF8}{gbsn}XX  +公司\end{CJK}" and is recognized as organization type by our Vanilla+Reg-aware model. After equipping with the regularity-agnostic module, we obtain the precise character boundary and relieve the excessive attention to regularity.

\section{Conclusion}
In this paper, we proposed a simple but effective method to explore the regularity information for Chinese NER, dubbed as Regularity-Inspired reCOgnition Network (RICON). It contains a regularity-aware module to capture the internal regularity feature of each span, and a regularity-agnostic module to reinforce the entity boundary detection while avoid imposing excessive attention on regularity. The features of two modules are encouraged to be dissimilar by an orthogonality space restriction. Evaluation shows that RICON achieves the state-of-the-art performance on four datasets.

\bibliographystyle{acl_natbib}
\bibliography{acl_latex}

\appendix

\section{Data Statistics}
\label{sec:appendix}
Table 6 shows the detailed statistics of each dataset.
\begin{table}[h]\small
\centering
\setlength{\tabcolsep}{1.45mm}{
\begin{tabular}{lcccc}
\hline
Datasets                        & Type     & Train   & Dev    & Test   \\ \hline \hline 
\multirow{3}{*}{OntoNotes V4.0} & Sentence & 15.7K   & 4.3K   & 4.3K   \\
                                & Char     & 491.9K  & 200.5K & 208.1K \\
                                & Entity   & 12.8K   & 6.5K   & 7.2K   \\ \hline
\multirow{3}{*}{OntoNotes V5.0} & Sentence & 36K    & 6.1K   & 4.5K   \\
                                & Char     & 1197.5K & 173.3K & 147.4K \\
                                & Entity   & 58.1K   & 8.5K   & 7.0K   \\ \hline
\multirow{3}{*}{MSRA}           & Sentence & 46.4K   & -      & 4.4K   \\
                                & Char     & 2169.9K & -      & 172.6K \\
                                & Entity   & 69.7K   & -      & 5.2K   \\ \hline
\multirow{3}{*}{CBLUE-CMeEE}           & Sentence & 15.3K   & 5.0k      & -   \\
                                & Char     & 825.0K & 270.4K      & - \\
                                & Entity   & 62.0K   &  20.3K     & -   \\ \hline
\end{tabular}}
\caption{Statistics of datasets.}
\label{Tab1}
\end{table}

\renewcommand\arraystretch{1.2}
\begin{CJK}{UTF8}{gbsn}
\begin{table*}[h]\small
\centering
\begin{tabular}{|l|ccccccc|}
\hline
Sentence (Truncated)         & \multicolumn{7}{l|}{\begin{tabular}[c]{@{}l@{}}铼 德 \textcolor{orange}{和} 年 兴 纺 织， 是 台 湾 唯 二 上 榜 的 公 司。\\ (RITEK \textcolor{orange}{and} Nien Hsing Textiles are the only two companies on the list in Taiwan.)\end{tabular}} \\
Characters (Entity Included) & 铼                   & 德                   & \textcolor{orange}{和}                   & 年                   & 兴                   & 纺                   & 织                   \\
Gold Label              & B-ORG               & E-ORG               & O                   & B-ORG               & M-ORG               & M-ORG               & E-ORG               \\
Vanilla             & O                   & B-ORG               & M-ORG               & M-ORG               & M-ORG               & M-ORG               & E-ORG               \\
Vanilla + Reg-aware              & B-ORG               & E-ORG              & O                   & B-ORG               & M-ORG               & M-ORG               & E-ORG               \\
Regularity weight             & 7.5e-2               & 2.5e-6               & \textbf{9.2e-1}              & 2.1e-6               &   8.9e-6                  &     9.5e-4                &       3.1e-5              \\

\hline                                           
\end{tabular}
\caption{An example on the Ontonotes V4.0 dataset. The label is organized in the form of BMES.}
\label{Tab5}
\end{table*}
\end{CJK}

\renewcommand\arraystretch{2}
\begin{table*}[t]\scriptsize
\centering
\setlength{\tabcolsep}{1 mm}{
\begin{tabular}{|llll|l|}
\hline
Sentence (Truncated)                                                     & Golden Entity / Type           & Biaffine Prediction               & RICON Prediction          & Regularity                        \\ \hline
  \begin{CJK}{UTF8}{gbsn}(1) 肺多叶病变显示...\end{CJK}                                                        &\begin{CJK}{UTF8}{gbsn}肺多叶病变，Symptom\end{CJK}& \begin{CJK}{UTF8}{gbsn}\textcolor{red}{未识别}\end{CJK}                         & \begin{CJK}{UTF8}{gbsn}肺多叶病变，Symptom\end{CJK}  & \multirow{3}{*}{\begin{CJK}{UTF8}{gbsn}XX+病变\end{CJK}}        \\
Multi-lobed lung lesions showed that...             & Multi-lobed lung lesions  &  \textcolor{red}{N/A}                          & Multi-lobed lung lesions &                                  \\ \cline{1-4}
\begin{CJK}{UTF8}{gbsn}(2) 大片状融合性病变为主...\end{CJK}                                                         & \begin{CJK}{UTF8}{gbsn}大片状融合性病变，Symptom \end{CJK} &\begin{CJK}{UTF8}{gbsn}\textcolor{red}{未识别}\end{CJK}                          & \begin{CJK}{UTF8}{gbsn}大片状融合性病变，Symptom\end{CJK}  &                                   \\
Massive fusion lesions were the main...              & Massive fusion lesion   &\textcolor{red}{N/A}    & Massive fusion lesion                      &  \multirow{3}{*}{XX+lesion} \\ \cline{1-4}
\begin{CJK}{UTF8}{gbsn}(3) 增加肝脏病变...多数属...\end{CJK}                                                     & \begin{CJK}{UTF8}{gbsn}肝脏病变，Symptom\end{CJK}&\begin{CJK}{UTF8}{gbsn}\textcolor{red}{未识别}\end{CJK}                          & \begin{CJK}{UTF8}{gbsn}肝脏病变，Symptom\end{CJK}&                                   \\
Increase liver lesions, most of which...                   & liver lesions &\textcolor{red}{N/A}                          & liver lesions           &                                   \\ \hline
\begin{CJK}{UTF8}{gbsn}(4) ...患儿可并发肝损害。\end{CJK}                                                          & \begin{CJK}{UTF8}{gbsn}肝损害，Symptom\end{CJK}                      & \begin{CJK}{UTF8}{gbsn}肝损害，\textcolor{red}{Disease}\end{CJK}&\begin{CJK}{UTF8}{gbsn}肝损害，Symptom\end{CJK} & \multirow{3}{*}{ \begin{CJK}{UTF8}{gbsn}XX+损害\end{CJK}}        \\
...can be complicated with liver damage.             & liver damage          & liver damage            &  liver damage   &                           \\ \cline{1-4}
\begin{CJK}{UTF8}{gbsn}(5)  SARS患儿有部分出现心脏损害...\end{CJK}                                          & \begin{CJK}{UTF8}{gbsn}心脏损害，Symptom\end{CJK}                     & \begin{CJK}{UTF8}{gbsn}心脏损害，\textcolor{red}{Disease}\end{CJK} & \begin{CJK}{UTF8}{gbsn}心脏损害，Symptom\end{CJK} &                                 \\
SARS children suffer from heart damage...            & heart damage          & heart damage  & heart damage         & \multirow{3}{*}{XX+damage} \\ \cline{1-4}
\begin{CJK}{UTF8}{gbsn}(6) ...合并多脏器损害\end{CJK}                                            & \begin{CJK}{UTF8}{gbsn}多脏器损害，Symptom\end{CJK}                    & \begin{CJK}{UTF8}{gbsn}多脏器损害，\textcolor{red}{Disease}\end{CJK}&    \begin{CJK}{UTF8}{gbsn}多脏器损害，Symptom\end{CJK}                   &                                   \\
...complicated with multi-organ damage      & multi-organ damage & multi-organ damage & multi-organ damage  &                                   \\ \hline
\end{tabular}}
\caption{Cases study on the domain CBLUE-CMeEE dataset.}
\end{table*}

\section{More Case Study}
\label{sec:appendix}

\subsection{Complex Regularity}
Besides explicit patterns like the first example in Table 5, Table 7 shows a more complex form of regularity that our model can capture.
In this example, the Vanilla+Reg-aware model pays highest attention weight 0.92 to important character ”\begin{CJK}{UTF8}{gbsn}和\end{CJK}” (and), and recognize that A and B are independent entities according to the regularity “\begin{CJK}{UTF8}{gbsn}A 和 B\end{CJK} (A and B)”. For comparison, the vanilla fails to distinguish these two entities. This example further reveals that our regularity-aware module can discover more complex character compositions.

\subsection{Case Study in Medical Domain}
To further demonstrate the effectiveness of our RICON in Chinese NER, we present six examples of the CBLUE-CMeEE dataset from the medical domain. 
As shown in the first three examples in Table 8, the biaffine  model fails to identify the accurate boundary of the entities, thus leading to unrecognized entity type. However, our RICON achieves detecting the correct span boundary as well as predicting golden type type (Symptom) of the entities according to the regularity "\begin{CJK}{UTF8}{gbsn}XX+病变\end{CJK}" (XX+lesion). 
In the last three examples, both biaffine model and our RICON successfully detect the correct span boundary of the entities. For entity type prediction, the biaffine model assigns a wrong type (Disease) to these entities, but our RICON predicts types correctly as a result of it captures the regularity feature "\begin{CJK}{UTF8}{gbsn}XX+损害\end{CJK}" (XX+damage) from "Symptom" type. To sum up, our RICON is also beneficial for domain datasets.

\end{document}